\documentclass{article} 
\usepackage{iclr2025_conference,times}

\usepackage{amsmath,amsfonts,bm}

\def\eqref#1{equation~\ref{#1}}

\def\1{\bm{1}}

\DeclareMathAlphabet{\mathsfit}{\encodingdefault}{\sfdefault}{m}{sl}
\SetMathAlphabet{\mathsfit}{bold}{\encodingdefault}{\sfdefault}{bx}{n}

\usepackage{lipsum}
\usepackage{graphicx}
\usepackage{hyperref}
\usepackage{url}

\graphicspath{ {./images/} }

\usepackage{amsthm}
\usepackage{amsmath}
\usepackage{mathtools}
\usepackage{graphicx}
\usepackage{natbib} 
\usepackage{amssymb}
\usepackage{booktabs}
\usepackage{hyperref}
\usepackage{tikz}
\usepackage{cleveref}
\usetikzlibrary{shapes, positioning}
\usepackage{wrapfig}

\newtheorem{example}{Example}

\newcommand{\Prob}{\mathbb{P}}

\newlength{\nodesep}
\newlength{\innersep}
\newlength{\sep}
\tikzset{
    observed/.style={circle, draw, thick, fill=gray!50, minimum size=.7cm},
    unobserved/.style={circle, draw, thick, minimum size=1.2cm},
    blank/.style={circle, draw, thick, dotted, minimum size=.7cm, opacity=0.5},
    plate/.style={draw, dotted},
    square1/.style={draw, fill=gray!20, minimum size=.3cm},
    square2/.style={draw, fill=gray!50, minimum size=.3cm},
    square3/.style={draw, fill=gray!100, minimum size=.3cm},
    true/.style = {circle, draw=none, thick, fill=green!50, minimum width=.5cm, minimum height=.5cm, inner sep=0pt},
    false/.style = {circle, draw=none, thick, fill=red!50, minimum width=.5cm, minimum height=.5cm, inner sep=0pt},
    plain/.style = {circle, draw, thick, fill=none, minimum width=.5cm, minimum height=.5cm, inner sep=0pt},
    arrowstyle/.style={->, thick, rounded corners},
    arrowstyledash/.style={->, thick, rounded corners, dash pattern=on 2pt off .6pt, gray},
    adjweight/.style={->, dashed, line width=.1mm, color=blue!50},
    adjweightbig/.style={->, line width=.6mm, color=blue!50},
    cembtrue1/.style={draw, fill=green!20, minimum size=.3cm},
    cembtrue2/.style={draw, fill=green!50, minimum size=.3cm},
    cembtrue3/.style={draw, fill=green!100, minimum size=.3cm},
    cembfalse1/.style={draw, fill=red!20, minimum size=.3cm},
    cembfalse2/.style={draw, fill=red!50, minimum size=.3cm},
    cembfalse3/.style={draw, fill=red!100, minimum size=.3cm},
    operation/.style={draw, thick, fill=blue!20, minimum size=0.6cm,  rounded corners=5pt, inner sep=1pt, outer sep=0pt, align=center},
}

\newcommand{\equivariance}{
\begin{tikzpicture}[
    node distance = .3cm,
]
\node[] (intH) at (0,0) {$V^{(m)}$};
\node[] (intM) at (2.5*\sep,0) {$\mathbb{P}_{V^{(m)}}^{a(i)}$};
\node[] (PH) at (0,-1.5*\sep) {$V^{(h)}$};
\node[] (PM) at (2.5*\sep,-1.5*\sep) {$\mathbb{P}_{V^{(h)}}^{a(\omega(i))}$};

\draw[arrowstyle] (intH) -- (PH) node[midway, left] {$\tau$};
\draw[arrowstyle] (intH) -- (intM) node[midway, above] {$a(i)$};
\draw[arrowstyle] (intM) -- (PM) node[midway, right] {$\mathcal{\tau}$};
\draw[arrowstyle] (PH) -- (PM) node[midway, below] {$a(\omega(i))$};

\end{tikzpicture}
}

\newcommand{\equivarianceSurrogate}{
\begin{tikzpicture}[
    node distance = .3cm,
]
\node[] (intH) at (0,0) {$V^{(m)}$};
\node[] (intM) at (3*\sep,0) {$\mathbb{P}_{V^{(m)}}^{a(i)}$};
\node[] (PH) at (0,-1.5*\sep) {$V^{(s)}$};
\node[] (PM) at (3*\sep,-1.5*\sep) {$\mathbb{P}_{V^{(s)}}^{a(\omega(i))}$};
\node[] (PS) at (0,-3*\sep) {$V^{(h)}$};
\node[] (PS2) at (3*\sep,-3*\sep) {$\mathbb{P}_{V^{(h)}}^{a(\omega'(\omega(i)))}$};

\draw[arrowstyle] (intH) -- (PH) node[midway, left] {$\tau$};
\draw[arrowstyle] (intH) -- (intM) node[midway, above] {$a(i)$};
\draw[arrowstyle] (intM) -- (PM) node[midway, right] {$\tau$};
\draw[arrowstyle] (PH) -- (PM) node[midway, below] {$a(\omega(i))$};
\draw[arrowstyle] (PH) -- (PS) node[midway, right] {$\tau'$};
\draw[arrowstyle] (PM) -- (PS2) node[midway, right] {$\tau'$};
\draw[arrowstyle] (PS) -- (PS2) node[midway, below] {$a(\omega'(\omega(i)))$};

\end{tikzpicture}
}

\title{Neural Interpretable Reasoning}

\makeatletter
\renewcommand\@fnsymbol[1]{%
  \ensuremath{%
    \ifcase#1
      \or
      \dagger 
      \or
      \ddagger 
      \else
      \@ctrerr
    \fi
  }
}
\makeatother

\author{
Pietro Barbiero$^*$ \\
IBM Research, Switzerland\thanks{Work conducted while employed at Università della Svizzera italiana.} \\
\texttt{pietro.barbiero@ibm.com} \\
\And
Giuseppe Marra$^*$ \\
KU Leuven, Belgium \\
\texttt{giuseppe.marra@kuleuven.com} \\
\And
Gabriele Ciravegna \\
Politecnico di Torino, Italy \\
\And
David Debot \\
KU Leuven, Belgium \\
\And
Francesco De Santis \\
Politecnico di Torino, Italy \\
\And
Michelangelo Diligenti \\
Universita' di Siena, Italy \\
\And
Mateo Espinosa Zarlenga \\
University of Cambridge, UK \\
\And
Francesco Giannini \\
Scuola Normale Superiore, Italy
}

\iclrfinalcopy 
\begin{document}

\maketitle

\begin{abstract}
We formalize a novel modeling framework for achieving interpretability in deep learning, anchored in the principle of inference equivariance. While the direct verification of interpretability scales exponentially with the number of variables of the system, we show that this complexity can be mitigated by treating interpretability as a Markovian property and employing neural re-parametrization techniques. Building on these insights, we propose a new modeling paradigm---\emph{neural generation and interpretable execution}---that enables scalable verification of equivariance. 
This paradigm provides a general approach for designing Neural Interpretable Reasoners that are not only expressive but also transparent.

\end{abstract}


\section{A Turing test for interpretability}

Interpretability, much like intelligence, is often subject to debate due to its inherently subjective nature~\citep{kim2016examples,miller2019explanation,molnar2020interpretable}. Instead of attempting to provide an exhaustive definition, in this paper we propose a procedural test---akin to the Turing test~\citep{turing1950computing}---that evaluates whether a system is interpretable.
We motivate our proposal using the following concrete examples.
\begin{example}
Donald Duck attempts to start his car, model 313, but the vehicle fails to start. After inspecting the situation, he finds that the fuel level is too low. Once he refuels, the car starts without issue. In this instance, Donald clearly understands the problem and its straightforward solution.
The following day, the car fails to start once more despite having a full fuel tank. Uncertain of the cause, Donald consults a mechanic. Building on her expertise in engines, the mechanic determines that an oil leak is the root of the problem. After repairing the leak, the car operates normally. Here, while Donald could not diagnose the issue on his own, his recourse to expert knowledge ultimately resolved the problem.
\end{example}
These examples illustrate that understanding a system is often subjective and dependent on the user's background~\citep{miller2019explanation}. However, they also suggest a practical criterion to check whether a system is interpretable. We can informally describe this criterion as follows:
\begin{quote}
\emph{A system is interpretable to a user if the user is able to interact with it and accurately forecast the system outputs.}
\end{quote}
This approach emphasizes the role of user interaction in assessing interpretability and mirrors the spirit of the Turing test by focusing on the system behavior. 

\paragraph{Contributions} This work's purpose can be characterized as threefold:
\begin{itemize}
    \item \textbf{Formalize interpretability as inference equivariance:} We formalize interpretability as\textit{ human-machine inference equivariance} and show that verifying inference equivariance directly is intractable (Sec.~\ref{sec:equiv}).
    \item \textbf{Break combinatorial complexity in verifying interpretability:} We show how the combinatorial complexity in verifying inference equivariance can be mitigated considering interpretability as a Markovian property and using techniques such as neural re-parametrization and mixture models (Sec.~\ref{sec:scalability}).
    \item \textbf{Formalize a modeling paradigm guaranteeing expressivity and interpretability by design:} Building on these insights, we propose a new modeling paradigm---\emph{neural generation, interpretable execution}---that enables scalable verification of interpretability and designing models that are not only expressive but also transparent (Sec.~\ref{sec:nir}).
\end{itemize}



\section{Interpretability \& equivariance}\label{sec:equiv}
\begin{wrapfigure}{r}{0.5\linewidth}
    \centering
    \includegraphics[width=\linewidth]{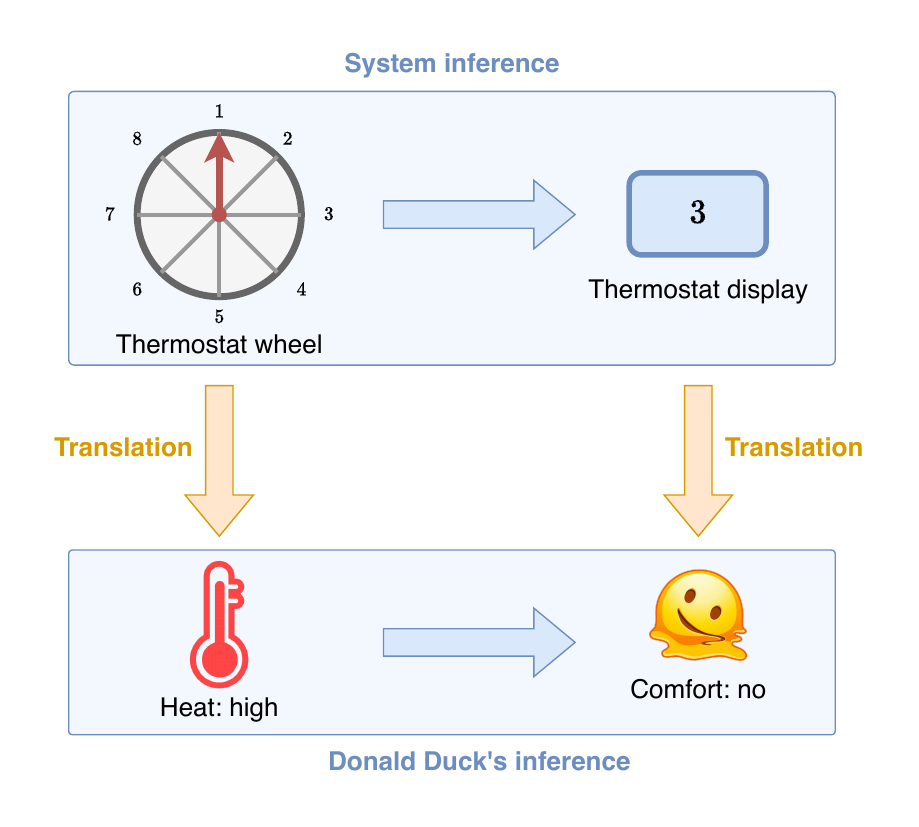}
    \caption{Example of inference equivariance.}
    \label{fig:equiv_example}
\end{wrapfigure}
Our work is motivated by the idea that a system is interpretable if its internal processes can be reliably translated into outcomes that users can predict. 
In this section, we formalize this notion as \textit{interpretability equivariance}, establishing that performing inference using the system's mechanisms should commute with the process of inference performed using the user's mechanisms.  We begin by motivating and illustrating this definition via an example:
\begin{example}[The Donald Duck Comfort Problem (Fig.~\ref{fig:equiv_example})]
Donald Duck wants to sleep but is uncomfortably cold. To achieve a comfortable sleep, he needs to warm up his environment to an appropriate temperature. A thermostat, whose user's manual Donald misplaced, controls the heating system. The thermostat provides only two pieces of information: a wheel with eight positions (currently set to $1$) and a numeric display ranging from 0 to 10 (currently showing $3$). 

In his first attempt, Donald rotates the wheel to position $6$. After waiting, he returns to observe that the display now reads $1$, and he finds himself sweating and uncomfortable. Donald can explain the phenomenon along two equivalent reasoning paths:
\begin{align*}
    \text{Thermostat path:} & \quad \texttt{wheel}=6 \to \texttt{display}=1 \to \texttt{comfort}= \text{no}, \\
    \text{Donald Duck path:} & \quad \texttt{wheel}=6 \to \texttt{heat}=\text{high} \to \texttt{comfort}= \text{no}.
\end{align*}
From this, Donald infers that turning the wheel upward increases the room’s temperature and causes the display to show lower numbers. To test his hypothesis, he sets the wheel to position $4$. Later, he checks the thermostat to find that the display now shows $2$, and he expects the room to have cooled down enough to restore his comfort:
\begin{align*}
    \text{Thermostat path:} & \quad \texttt{wheel}=4 \to \texttt{display}=2 \to \texttt{comfort}= \text{yes}, \\
    \text{Donald Duck path:} & \quad \texttt{wheel}=4 \to \texttt{heat}=\text{medium} \to \texttt{comfort}= \text{yes}.
\end{align*}
\end{example}

The example illustrates that while the thermostat's variables differ in semantics from Donald Duck's internal concepts, they are nonetheless aligned closely enough for him to establish a straightforward mapping between the two. For instance, a wheel position within the range $[3, 4]$ might be interpreted as medium heat, and a display reading of $2$ may be associated with a state of comfort. Furthermore, Donald's reasoning demonstrates that he can deduce the system's state via two equivalent routes--either by consulting the display or by directly sensing the heat output--with both methods leading to the same conclusion. Building on this intuition, we first introduce some useful notation and then use this to formalize our notion of interpretability equivariance. 

\subsection{Interpretability as inference equivariance}
\paragraph{Preliminaries: Transformation of Random Variables} 
Let $V$ denote a set of random variables representing different aspects of a system (for example, heating levels, wheel position, etc.). We write the joint probability distribution of these variables as 
$
\Prob(V) = \Prob(V_1, V_2, \cdots, V_n).
$
To formalize the distinction between the internal, machine-oriented description of the system and its human-interpretable counterpart, we index machine-related variables with the superscript $m$ (so that $V^{(m)}$ represents the machine’s variables) and human-related variables with $h$. Following~\citet{rubenstein2017causal}, we define a \emph{translation function} 
$
\tau: V^{(m)} \to V^{(h)}
$
as a map between machine variables and the variables within the human's reference system. Consequently, for any distribution $\Prob(V^{(m)})$ over the machine variables, the corresponding distribution in the human space is given by the push-forward measure $\Prob_{\tau(V^{(m)})} = \tau\bigl(\Prob_{V^{(m)}}\bigr)$. In particular, for each action on the $i$-th machine variable $a(i)$ (e.g., observing $a(i) \coloneqq \bigl( V_i^{(m)}=k \bigr)$ or intervening on the value of a variable $a(i) \coloneqq do(V_i^{(m)})$), we can define
the induced distribution $\Prob_{\tau(V^{(m)})}^i = \tau\bigl(\Prob_{V^{(m)}}^{a(i)}\bigr)$. 
To exactly transform the machine system into the human system, we require a surjective mapping $\omega: I_{V^{(m)}} \to I_{V^{(h)}}$ that assigns machine variable indices to human variable indices such that $\mathbb{P}_{\tau(V^{(m)})}^i = \mathbb{P}_{V^{(h)}}^{a(\omega(i))}$. Rather than enforcing that $\omega$ be order-preserving as in \cite{rubenstein2017causal}, our formulation of \emph{inference equivariance} requires that $\omega$ preserves conditional independence relations (which represents a weaker requirement). Formally, define the neighborhood of a machine variable $V_i^{(m)}$ as the minimal set of variables rendering it conditionally independent of the rest, i.e., 
\[
\mathcal{N}(V_i^{(m)}) = \min \{\, S^{(m)} \subseteq V^{(m)} \setminus \{V_i^{(m)}\} : V_i^{(m)} \perp (V^{(m)} \setminus (\{V_i^{(m)}\} \cup S^{(m)})) \mid S^{(m)} \,\}.
\]
We say that $\omega$ preserves conditional independencies if and only if, for every $V_i^{(m)}$ and every subset $S^{(m)} \subseteq V^{(m)} \setminus \{V_i^{(m)}\}$, 
\[
V_i^{(m)} \perp (V^{(m)} \setminus (\{V_i^{(m)}\} \cup S^{(m)})) \mid S^{(m)}
\]
if and only if 
\[
\tau(V_{\omega(i)}^{(m)}) \perp ( \tau(V^{(m)}) \setminus (\{ \tau(V_{\omega(i)}^{(m)})\} \cup \tau(S^{(m)}))) \mid \tau(S^{(m)}).
\]
This condition ensures that $\omega$ precisely mirrors the conditional independence structure between the machine and human systems.


\paragraph{Inference equivariance} 
The principle of \textit{inference equivariance}, illustrated in our previous example, asserts that the process of translating a machine's probability distribution into the human reference system and then querying it should yield the same result as first performing the query within the machine's domain and then translating the result. Formally, this is expressed as
\[
\resizebox{0.3\textwidth}{!}{\equivariance}
\]
This equality encapsulates the idea that whether one chooses to ``translate, then query" or to ``query, then translate", the resulting inference remains the same, as already observed for causal structures~\citep{rubenstein2017causal,geiger2024causal,marconato2023interpretability}. In the context of the Donald Duck example, this principle becomes particularly clear. Donald Duck faces a thermostat whose internal variables---such as the wheel setting and display reading---are not immediately aligned with his intuitive notions of heat and comfort\footnote{Notice that in contrast with equivariances in causal abstractions~\citep{geiger2024causal} where the inference structure is assumed to be aligned with the true data generating mechanisms.}. By establishing a mapping between the machine's outputs and his own reference system, he is able to reliably predict his comfort level. 

For instance, Donald might first translate the thermostat's raw signal (the display reading) into his internal concept of temperature and then infer his comfort state based on that interpretation. Alternatively, he might directly observe the mechanical behavior (the wheel position) to predict the corresponding change in room temperature, and only afterwards translate that information into his subjective experience of warmth. The fact that both routes lead him to the same conclusion---whether he ``translates, then queries" or ``queries, then translates"---demonstrates the principle of inference equivariance. 

This consistency is critical: it ensures that the mapping between machine variables and human concepts is robust, thereby making the system interpretable. In essence, the equality \emph{``translate, then query'' = ``query, then translate''} guarantees that a user's understanding and predictions of a system's behavior remain coherent, regardless of the order in which translation and inference occur.

\paragraph{Verify interpretability via inference equivariance is intractable} 
While the concept of equivariance provides a robust framework for linking machine and human perspectives, its practical implementation is fraught with challenges. As the number of variables increases, verifying and maintaining equivariance becomes exponentially more complex. To illustrate, consider a simple scenario where every variable in the system is Boolean. In this simple case, a complete interpretation of the system would require verifying the equivariance for all possible states of the system. This corresponds to extracting the full conditional probability table, which contains $2^n$ entries for $n$ variables. Even for a modest $n$, the number of combinations quickly becomes computationally intractable.
For this reason, in practical applications it becomes essential to guarantee inference equivariance indirectly or approximately while maintaining computational efficiency. In practice, this may involve constraining the inference space to a subset of critical variables, leveraging problem-specific structures to reduce complexity, or employing surrogate models that approximate the full system's behavior with a significantly lower computational cost.

\subsection{Properties of interpretability through the lenses of inference equivariance}
Based on inference equivariance, we can highlight several key properties that further clarify the nature of interpretability.

\textbf{Inference equivariance can be asymmetric:} In the thermostat example, Donald Duck uses the available signals---such as the wheel position and the display reading---to form an understanding of the system's behavior. Importantly, for him to use the thermostat effectively, it is unnecessary to have a complete, invertible mapping from his internal concepts (e.g., ``comfort level'') back to the machine's variables. This one-way, asymmetric mapping suffices because Donald only needs to translate machine outputs into human-understandable signals. The absence of a reverse transformation does not impede his ability to predict the system's response, illustrating that the forward mapping (machine $\rightarrow$ human) is all we require for interpretability (although the opposite mapping might be needed for supervised learning).

\textbf{Explanations are a form of selection:} An explanation of a system's behavior can be seen as a process of selection, where conditioning on observed evidence picks out a specific subset from the system's complete conditional probability table. In the Donald Duck example, when Donald observes a particular display reading or wheel position, he effectively selects a corresponding segment of the conditional probability table that relates these inputs to his comfort state. This selection---formally represented with the distribution $\Prob(V \mid a(V'))$---encapsulates the explanation by narrowing down the myriad potential outcomes to the ones relevant to his observation.

\textbf{Explanations might not be interpretable:} Not every selection from the conditional probability table yields a meaningful or interpretable explanation. For example, if the mapping between the thermostat's signals and Donald's perception of warmth were inconsistent---if the transformation did not commute---then the same action might lead to different inferred comfort states, confusing the user. Hence, for an explanation to be interpretable, the diagram representing the transformation must commute, ensuring that no matter how the inference is performed, the resulting explanation is consistent and understandable.


\textbf{Local vs. global equivariance:} Equivariance may hold over the entire state space of the system (global) or only in certain regions (local). In the case of the thermostat, Donald Duck might have developed an accurate translation for a subset of wheel positions, while other settings remain ambiguous. This local equivariance indicates that while the system may be interpretable under specific conditions, \textit{its interpretability might not generalize across all possible configurations}. Recognizing the distinction between local and global equivariance is crucial for assessing the robustness of a system's interpretability.

\textbf{Post-hoc methods complicate rather than simplify interpretability:} When applying post-hoc interpretability techniques, such as using surrogate models to explain the original system~\citep{hinton2015distilling, deepred} or so-called feature importance methods~\citep{lime, shap, og_saliency, integrated_gradients}, an additional layer of equivariance is required. Suppose Donald employs a surrogate model to better understand his thermostat. In that case, there must be a consistent mapping between the machine variables of the original system $V^{(m)}$ and those of the surrogate model $V^{(s)}$ and another mapping from the surrogate model to Donald Duck $V^{(h)}$. Formally, 
both the original and surrogate systems must satisfy the inference equivariance conditions:
\[
\resizebox{0.3\textwidth}{!}{\equivarianceSurrogate}
\]
This requirement ensures that the explanations generated by the surrogate model faithfully reflect the behavior of the original system, thus preserving interpretability even when using post-hoc methods. Ultimately, the need to establish these additional mappings significantly complicates the interpretability process as two equivariance relations must be satisfied instead of one.

\subsection{Semantic and functional equivariances}

Previous works~\citep{geiger2024causal,marconato2023interpretability} focused primarily on semantic equivariance, emphasizing that equivariance should hold for random variables $V$. However, less attention has been paid to the functions that describe the mappings between random variables; for a user to truly understand the underlying mechanisms, the structure of the function and its parameters must also satisfy equivariance, as illustrated in the following example.
\begin{example}
Consider the conditional model $\Prob(V_2 \mid V_1)$ where $V_2$ follows a Gaussian distribution:
\[
\Prob(V_2 = v; \mu=V_1, \sigma) \coloneqq \frac{1}{\sqrt{2\pi\sigma^2}} \exp\!\left(-\frac{(v-\mu)^2}{2\sigma^2}\right).
\]
For this model to be fully interpretable, it is not enough for a human user to simply understand the data representation encoded in $V_1$ and $V_2$. Instead, inference equivariance must extend to the functional structure and its parameters. In other words, users should be able to modify or update the parameters---such as $\mu$ or $\sigma$, or even alter constants like replacing $2\pi$ with $3\pi$---and still verify that the same equivariant relations hold. This ensures that the underlying functional form of the model remains transparent.
\end{example}

The intuition behind this is that functional structure and parameters are key components of interpretability, not just the data representations. To capture this formally, we can distinguish between variables representing data, $V \in \mathcal{V}$, and those describing the model's functional structure, $\theta \in \Theta$. The complete model can then be expressed as $\Prob(V, \Theta)$. Inference equivariance should hold for both $V$, ensuring \textit{semantic transparency}, and for $\theta$, ensuring \textit{functional transparency}.

\section{Breaking combinatorial complexity in verifying interpretability} \label{sec:scalability}
As we discussed verifying inference equivariance directly is intractable. In this section we discuss interpretability properties and techniques which can be used to break this complexity down.

\subsection{Interpretability is a Markovian property}
In the earlier thermostat example, Donald Duck successfully built an intuitive understanding of how the thermostat worked, despite having no specialized knowledge of electronics or physics. This observation illustrates how interpretability is a Markovian property: a user can interpret a system at a given level of abstraction without needing to reference lower-level details.
In this context, interpretability is achieved locally---each step of an inference process can be understood in isolation from others. We can formalize this Markovian property of interpretability by writing:
\begin{equation}    
    \forall \; V_i, V 
    \; \Prob \models (V_i \bot V ) \mid \mathcal{N}(V_i)
\end{equation}
meaning that, given its 1-hop neighborhood $\mathcal{N}(V_i)$, any variable $V_i$ is conditionally independent of all other variables. This property allows a user to interpret a single step of the inference process---the one concerning the variable $V_i$---without needing to backtrack through the entire chain of reasoning.

This Markovian property of interpretability attenuates scalability issues, as it permits the analysis of individual steps without the burden of interpreting the entire system at once. This layered approach is reflected in models such as Self-Explaining Neural Networks~\citep{alvarez2018towards}, Concept Bottleneck Models~\citep{koh2020concept}, 
or Prototypical Networks~\citep{chen2019looks}, where semantically interpretable components (e.g., the concept bottleneck) are designed to be interpretable on their own, regardless of previous layers. In the Donald Duck example, his ability to understand the thermostat's behavior without the need to understand its engineering shows the practical benefits of this Markovian property.

\subsection{Re-parametrizations break equivariance complexity while guaranteeing expressivity and interpretability}

Interpreting complex systems often entails dealing with a vast number of variables, which can overwhelm human cognitive limits:
\begin{example}[Thermostat with Many Knobs]
Consider a new thermostat design featuring 100 knobs, where a certain (unknown) set of knobs controls the room temperature for a given day of the calendar year. In this scenario, Donald Duck would need to test every possible knob configuration to fully understand how the thermostat works.
\end{example}
This example highlights a fundamental scalability issue: while a machine can, in principle, process and manage a large number of independent variables, human users typically can only handle around 7 $\pm$ 2 variables at any one time~\citep{miller1956magical}. It clear that even under the assumption that variables operate independently (which is quite common in the field of eXplainable Artificial Intelligence, or \textit{XAI}), the number of interactions required to understand the system grows linearly with the number of variables. For humans, who are limited to processing a constant number of variables simultaneously (i.e., 7 ± 2), this poses a significant obstacle to interpretability. The key question then becomes: how can we design a system that presents only a constant number of variables to a human, without sacrificing the system's overall expressivity?
A promising approach to manage this challenge is re-parametrization, where a system is transformed into an equivalent form that preserves its expressivity while reducing the number of variables a human must directly consider.

\paragraph{Functional Mixtures}
One effective strategy is to decompose a complex system into a mixture of simpler subsystems, each of which is easy to understand~\citep{mclachlan1988mixture}. For instance, imagine a thermostat with 365 knobs (so, even more than the original 100 knobs!), but with the twist that only one knob is active per day, and an indicator light signals which knob is relevant at that time. This design ensures that, at any given moment, Donald needs to focus on only one knob rather than hundreds. Such re-parametrization retains the full expressive power of the original system while offering local representations that are much more interpretable. Techniques like Self-Explaining Neural Networks~\citep{alvarez2018towards}, ProtopNets~\citep{chen2019looks}, and Concept Memory Reasoning~\citep{debot2024interpretable} embody this approach by generating simple, locally faithful explanations whose composition may form arbitrarily non-linear decision boundaries.

\paragraph{Functional and semantic re-parametrizations}
In many classification problems, re-parametrization involves two key components: mapping raw variables to higher-level concepts (\textit{semantic re-parametrization}) and decomposing complex function parameters into simpler mixtures (\textit{functional re-parametrization}). In this framework, the original data variables are transformed into a set of human-interpretable concepts, ensuring semantic transparency as in Concept Bottleneck Models~\citep{koh2020concept}. Simultaneously, the function that governs the model's behavior is restructured into a mixture of simple functions, which preserves the model's expressivity while making it easier to understand as in Self-Explaining Neural Networks~\citep{alvarez2018towards} and Concept Memory Reasoning~\citep{debot2024interpretable}.




\section{Neural Interpretable Reasoning} \label{sec:nir}
Building on our previous discussions of interpretability properties and leveraging techniques such as re-parametrizations, we propose a new modeling paradigm that guarantees the scalable verification of interpretability as inference equivariance. In this framework, the following elements are essential:
\begin{itemize}
    \item \textbf{Semantic transparency:} The model must employ high-level, human-understandable concepts (e.g., as in~\citet{tcav, koh2020concept, concept_whitening}).
    \item \textbf{Functional transparency:} The function that maps these concepts to the desired tasks should have a low-complexity structure (e.g., linear), and its parameters should be interpretable.
    \item \textbf{Markovian property of interpretability:} By focusing on a single layer of the system (for instance, the final classification layer), this approach breaks down the complexity that arises from having to interpret the concept generation (which requires a separate verification procedure).
    \item \textbf{Functional mixtures:} When working in a setup where there is a high number of concepts, \textit{functional mixtures} help manage the model's complexity by decomposing the mapping from concepts to tasks into simpler, more interpretable components.
    \item \textbf{Neural re-parametrizations:} Both concepts and functions can be neurally re-parametrized, allowing one to retain the model's expressivity after re-parameterization.
\end{itemize}
Together, these properties form the basis of a new modeling paradigm we refer to as \emph{neural generation and interpretable execution}, which ensures that interpretability equivariance can be verified in a scalable manner.

\paragraph{Neural Generation, Interpretable Execution}
To concretely instantiate our proposal, consider a classification problem where the objective is to predict a target label $Y$ from a set of low-level features (e.g., pixel intensities) $X$.
Rather than using an opaque monolithic model, we propose to leverage the expressive power of deep neural networks (DNNs) to generate (i) the parameters of a transparent model $W$, and (ii) human-understandable data representations $C$ (a.k.a., concepts)---which together form the elements of an interpretable system. The learned transparent model is then symbolically executed to make predictions $Y$:
\begin{equation}
    \Prob(Y \mid X; \theta) = \int_{W} \sum_C 
    \overbrace{\Prob(Y \mid C; W)}^{\shortstack{\scriptsize \text{interpretable execution} \\ \scriptsize  \textit{(interpretability)}}}  \overbrace{\Prob(C, W \mid x; \theta_g)}^{\shortstack{\scriptsize \text{neural generation} \\ \scriptsize  \textit{(accuracy)}}}
\end{equation}
These two factors represent the neural generation component $\Prob(C, W \mid X; \theta)$, which re-parametrizes concept representations and functional parameters to ensure expressivity, and the symbolic execution component $\Prob(Y \mid C; W)$, which guarantees interpretability in the decision-making process. 
We refer to the family of models implementing this paradigm as \textit{Neural Interpretable Reasoning}. This family integrates deep neural network expressivity with interpretability by combining semantic transparency, functional transparency, and scalable verification of inference equivariance. 
\begin{wrapfigure}[15]{r}{0.6\linewidth}
    \centering
    \includegraphics[width=\linewidth]{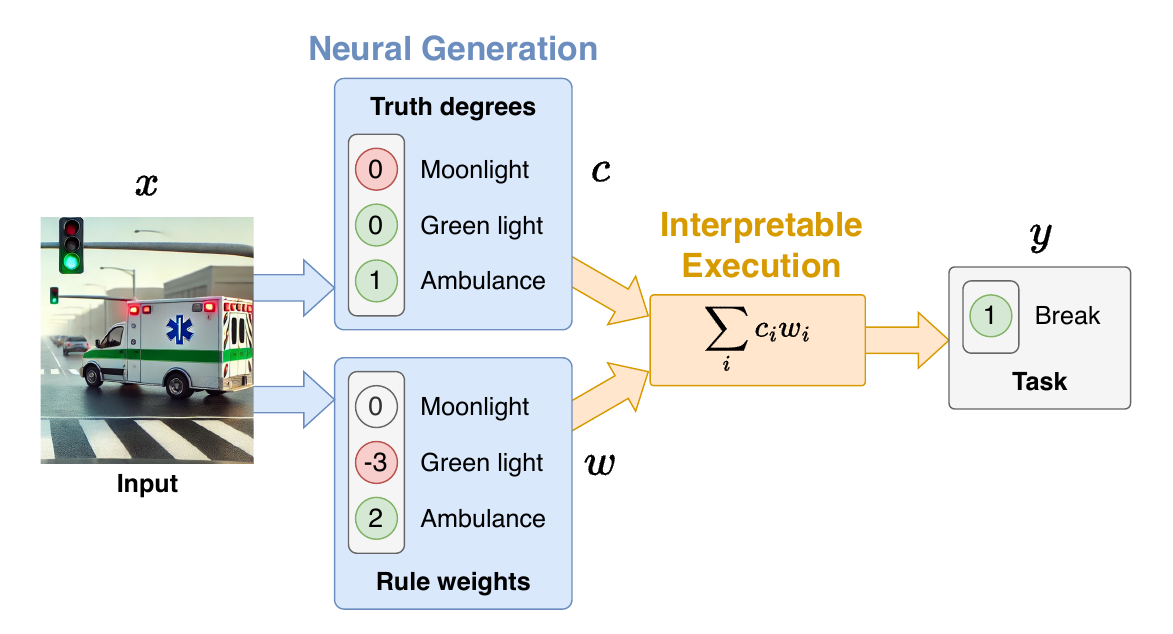}
    \caption{Neural Interpretable Reasoning.}
    \label{fig:nir}
\end{wrapfigure}
Fig~\ref{fig:nir} shows a NIR example where a self-driving car must decide whether to brake at an intersection. The architecture first generates both the truth degrees of relevant concepts (e.g., the presence of an ambulance or a green light) and the weights of a simple linear model (e.g., an ambulance is assigned a weight of $2$ because it is positively correlated with braking); then, the linear model is executed on these truth degrees to predict whether to brake.
Many well-known XAI techniques can be seen as special cases within this framework. For example, Prototypical Networks (ProtopNets)~\citep{chen2019looks}, Neural Additive Models~\citep{agarwal2021neural}, and Concept Bottleneck Models~\citep{koh2020concept} all embody aspects of interpretability that align with our proposed approach. More recently, novel approaches such as Concept Memory Reasoning~\citep{debot2024interpretable} and Explanation Bottleneck Models~\citep{yamaguchi2024explanation} have begun to fully exploit the potential of functional re-parametrization retaining the expressivity of traditional, opaque deep neural networks while supporting the scalable verification of interpretability.


\section{Conclusions}
In this paper, we introduced a novel framework for assessing and achieving interpretability, anchored in the principle of inference equivariance. Drawing inspiration from the Turing test procedure, we proposed that a system is interpretable if a user can reliably predict its behavior. In our discussion we argue that verifying interpretability directly scales exponentially in the number of variables even in simple cases. However, this complexity can be mitigated considering interpretability as a Markovian property and techniques such as neural re-parametrization which can break down complexity without sacrificing overall the model expressivity.
Building on these insights, we proposed a new modeling paradigm, \emph{neural generation and interpretable execution}, which integrates semantic transparency, functional transparency, and scalable verification of equivariance. This paradigm provides a promising pathway for designing Neural Interpretable Reasoners that are not only expressive but also transparent.

\bibliography{iclr2025_conference}
\bibliographystyle{iclr2025_conference}

\newpage
\appendix

\end{document}